# GENDER CLASSIFICATION USING GRADIENT DIRECTION PATTERN

**Mohammad Shahidul Islam**
Department of Computer Science, School of Applied Statistics,
National Institute of Development Administration, Bangkok, Thailand.
suva93@gmail.com

**ABSTRACT:** *A novel methodology for gender classification is presented in this paper. It extracts feature from local region of a face using gray color intensity difference. The facial area is divided into sub-regions and GDP histogram extracted from those regions are concatenated into a single vector to represent the face. The classification accuracy obtained by using support vector machine has outperformed all traditional feature descriptors for gender classification. It is evaluated on the images collected from FERET database and obtained very high accuracy.*

**Keywords:** Gender Classification, Gradient Direction Pattern, Texture Analysis, Pattern Recognition

## INTRODUCTION

It is very important to know the gender in social interactions. Human has a unique power to classify the gender from the view. Sometimes body movement also helps to identify the gender. Human face is a good source of information for identifying the person, facial expression, age and so on. Computer vision and pattern recognition is playing an important role for past few years. HCI (Human Computer Interaction) devices can be more user-friendly and act like human if it can extract gender information from human face [1] Gender information can also play a vital role in human-robot interaction; vision based human monitoring, passive demographic data collection, and human retrieval from video databases.

Image-based gender classification is difficult due to the inherent variability of human-face caused by age, ethnicity and image quality [2,3]. Features from facial region can be derived based on two methods [1]: appearance-based method and geometric-feature based method. Appearance-based methods consider the facial screen color, color difference, texture or color gradient direction to derive feature pattern from the image. It can be local region based e.g. local binary pattern [4], gradient direction pattern [5] or holistic e.g. Gabor Filter [6]. Geometric-based methods use location and distance between facial components like eyes, nose or mouth. Therefore, it needs extra computation before feature extraction to localize facial components. Any error during the component localization may lead to substantial accuracy drop.

Appearance-based methods get popularity due to its robustness in environmental change. Moreover local region-based feature extraction is independent to the location information of facial components, which makes it more favorite for the researchers. Local binary pattern is an example of local region-based feature descriptor, which is adopted by many researchers in the field of texture analysis for both object and human face. It was propose by Ojala *et al.* [4] for texture description, later used by Ahonen et al. [7] on human face for face recognition. An example of obtaining local binary pattern is shown in Figure 1.

| $S_1$ | $S_2$ | $S_3$ |
|---|---|---|
| $S_8$ | C | $S_4$ |
| $S_7$ | $S_6$ | $S_5$ |

$LBP(C) = \sum_{k=1}^{8} f(S_k, C) 2^{8-k}$, Where, $f(S,C) = \begin{cases} 0, & S < C \\ 1, & S \geq C \end{cases}$

And $S_{1-8}$ and C represents the gray value of that pixel

| 56 | 89 | 45 |   | 1 | 1 | 0 |
|---|---|---|---|---|---|---|
| 54 | 56 | 25 | → | 0 |   | 0 |
| 48 | 89 | 26 |   | 0 | 1 | 0 |

$LBP(56) = 1*2^7 + 1*2^6 + 0*2^5 + 0*2^4 + 0*2^3 + 1*2^2 + 0*2^1 + 0*2^0 = 196$

**Figure 1: Example of Local Binary Pattern**

Local binary pattern was also used for gender classification [8,9] and achieved very competitive results. LBP is gray-scale invariant as it uses the color difference to compute the binary pattern. Therefore, it is robust in illumination changes and performs better in an uncontrolled environment. However, its noise tolerance is very poor as little changes in some pixels intensity caused by noise changes the binary pattern. In addition, the LBP histogram length is too high, which is time and cost effective.

A new methodology Gradient Direction Pattern is proposed in this paper for gender recognition which was already proved to be cost and time effective as a facial feature descriptor for facial expression recognition [5]. In this paper, GDP is combined with Kirsch edge response mask 10 to make it more stable in the noise. Experiments were performed on FERET dataset [11] using LBP and GDP in matlab. Gaussian white noise was added to prove GDPs stability in noisy environment.

## FEATURE EXTRACTION

For each pixel in the gray-scale image, only surrounding eight pixels were used to extract feature. Firstly, that 3x3 pixels region is multiplied with Kirsch edge response mask in eight directions to get eight-mask value.

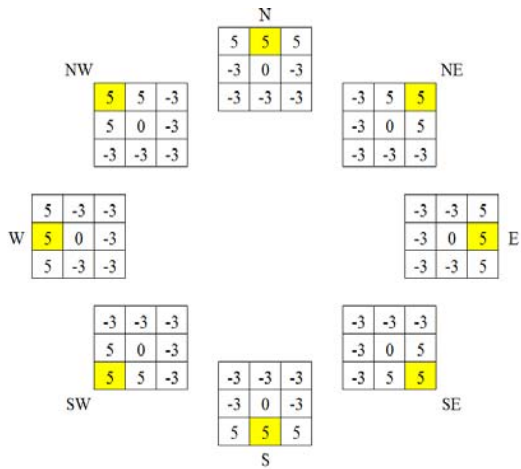

Figure 2: Kirsch edge response mask in eight directions.

After applying the directional masks, the pixels original color value is replaced by the corresponding mask value (Figure 3) and GDP is applied on that region to obtain GDP code (Figure 4) which is more stable in a noisy environment than general GDP code. GDP code is a 4-bit binary pattern, which can give up to 16 different combinations.

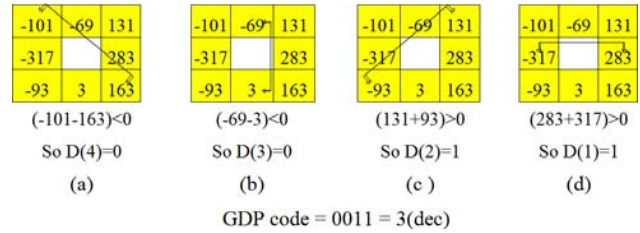

Figure 3: Obtaining GDP code from a 3x3 region

## FEATURE VECTOR

The gray scale image was divided into 81 equal sized blocks and histogram of GDP codes from each block is concatenated to form the feature vector (Figure 5). GDP considered at most single transition patterns only. Therefore, histogram length for each block was eight and the feature vector length for the whole image was 81*8= 648.

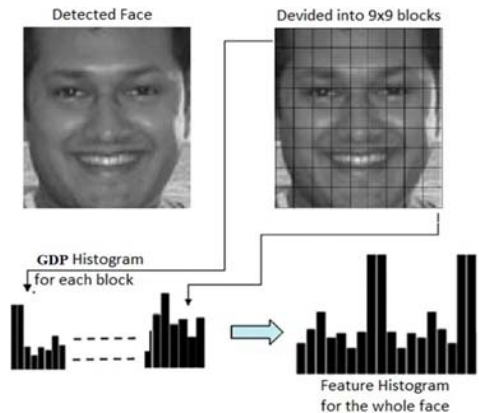

Figure 4: Obtaining mask value for a 3x3 region

Figure 5: Building feature vector for a gray scale image

## CLASSIFICATION USING SUPPORT VECTOR MACHINE

Support vector machine is a well known classier and is successfully used in many research work for classification [12]. It maps the feature data in to high dimensional feature space and draws a clear separation line between them. For linear data, it is easy to separate them but for non-linear data, SVM uses some sort of kernel function to create non-linear separation line. Some popular kernel functions are polynomial, RBF (radial basis function) etc. Support vector machine is a binary classifier.

## EXPERIMENTAL RESULTS

The performance of gender classification using proposed GDP feature is evaluated on FERET [11] database. The FERET database has a total of 14,051 gray-scale images from 1,199 subjects. The images have variations in lighting, facial expressions, pose angle, aging effects etc. For this study, 2000 mug-shot face images are collected out of which 1100 faces are male and rest 900 faces are female. Fdlibmex, a free face detecting tool for matlab is used to detect the facial area. After that each face is divided into NxN sub-blocks to generate spatially combined GDP histogram feature vector as discussed earlier. Experiments show that increasing the number of blocks can enhance the performance but it increases the length of the feature vector and requires extra bit of processing both time and space. The classification performance in respect to different number of blocks is shown with Table I. Performance of the proposed method in comparison other state of the art methods including LBP method is shown in Table II, which demonstrate superiority of proposed GDP feature over LBP and other feature in gender classification domain. Stability of GDP in noisy environment is shown in Table III.

**Table I: Classification Accuracy with Different Block.**

| Number of Block | Feature Length | Classification Accuracy | | |
|---|---|---|---|---|
| | | Overall | Male | Female |
| 5x5 | 80 | 95.15% | 95.31% | 94.99% |
| 7x7 | 392 | 94.94% | 94.65% | 95.23% |
| 9x9 | 648 | 95.96% | 95.94% | 95.98% |
| 11x11 | 968 | 95.96% | 95.91% | 96.01% |
| 13x13 | 1352 | 95.73% | 95.35% | 96.11% |

**Table II: Classification Accuracy with Other Method.**

| Methods | Classification Accuracy | | |
|---|---|---|---|
| (Feature + Classifier) | Overall | Male | Female |
| *GDP + SVM* | *95.96%* | *95.94%* | *95.98%* |
| LBP + Chi-square [13] | 81.90% | 82.27% | 81.44% |
| LBP + Adaboost [13] | 92.25% | 92.00% | 92.55% |
| LDP + SVM [14] | 95.05% | 94.81% | 95.33% |

**Table III:** Results of GDP+SVM and other popular methods in noisy environment

| Method | Classification Accuracy(Overall) | |
|---|---|---|
| | No noise | With white noise* |
| *GDP* | *95.96%* | *95.15%* |
| LBP | 91.68% | 87.41% |
| LBP$_U$ | 91.51% | 85.36% |

*Gaussian White noise of 0 mean and 0.001 variance is applied.

## CONCLUSION
A novel facial feature extraction method for gender classification is proposed in this paper named GDP (Gradient Direction Pattern). The GDP codes are insensitive to noise because it uses edge response value instead of the pixels color value to compute features. It considers only uniform patterns, which cuts the feature vector length by half. Therefore, GDP codes provide cost and time effective robust features to represent facial appearance. The classification accuracy achieved using powerful classifier SVM is very competitive. Future plan is to work with real time video.

## ACKNOWLEDGMENT
I am deeply grateful to my supervisor, Associate Professor Dr. Surapong Auwatanamongkol, Department of Computer Science, National Institute of Development Administration, Bangkok 10240, for his detailed and constructive comments, and for his important support throughout this work.